\documentclass{article} 
\pdfoutput=1
\usepackage{augie,times}
\usepackage{hyperref}
\usepackage{url}
\usepackage{graphicx}
\usepackage{amsmath,amsfonts,amssymb}
\usepackage{mathtools}
\usepackage{amsthm}
\usepackage{color}
\usepackage{algpseudocode}
\usepackage{algorithm}

\title{\scalebox{1}{Faster Asynchronous SGD}}

\author{
\hspace{14mm} Augustus Odena \\
\hspace{3mm} \texttt{augustus.odena@gmail.com}
}

\begin{document}

\begin{center}
\maketitle
\end{center}

\begin{abstract}

Asynchronous distributed stochastic gradient descent methods have trouble converging because of stale gradients.
A gradient update sent to a parameter server by a client is stale if the parameters used to calculate that gradient have since been updated on the server.
Approaches have been proposed to circumvent this problem that quantify staleness in terms of the number of elapsed updates.
In this work, we propose a novel method that quantifies staleness in terms of moving averages of gradient statistics.
We show that this method outperforms previous methods with respect to convergence speed and scalability to many clients.
We also discuss how an extension to this method can be used to dramatically reduce bandwidth costs
in a distributed training context.
In particular, our method allows reduction of total bandwidth usage by a factor of 5 with little impact on
cost convergence.
We also describe (and link to) a software library that we have used to simulate these algorithms
deterministically on a single machine.

\end{abstract}

\section{Introduction}

Neural Networks can be trained on multiple machines simultaneously using distributed stochastic gradient descent (SGD).
This can be done on commodity CPU clusters as in \cite{Dean} and \cite{ADAM},
or on a heterogenous mixture of machines as in \cite{ONEBIT} and \cite{Wu}.
In either case, a replica of the network is created on each separate machine. These replica machines are called clients.
One machine holds the canonical snapshot of the model parameters, and this machine is called the server.
Clients take a random mini-batch of training data and perform backpropagation on their replica of the model to maintain a gradient estimate.
This estimate is then sent back to the server to be applied to the global parameters.
If the server waits to collect all of the client updates before applying them to its parameters, we say that the SGD is synchronous.
If the server applies updates as they come in (as first introduced in \cite{FirstASGD}), we say that the SGD is asynchronous.
Each method has pros and cons.
Synchronous SGD is slower because clients spend some time waiting, but it's free from convergence issues.
Async SGD removes the waiting, but may have trouble converging due to stale gradients.
Thus, dealing effectively with stale gradients is essential to achieving good performance
for asynchronous distributed training algorithms when the number of clients is high.

In \cite{Rel}, they suggest an exponential penalty for stale gradients.
This could work for a small number of clients, but it will reduce the learning rate too far
when staleness values are large.

In \cite{Suyog} they propose a type of staleness aware async SGD.
In the context of asynchronous gradient updates in a parameter server context, they say a gradient
sent to the parameter server from a worker is stale by k steps if the weights used to compute that
gradient have since been updated k times.

They then describe a policy of modulating the learning rate as a function of that staleness measure.
This is accomplished by simply dividing the gradient update by the staleness measure before applying it.
We refer to that strategy as staleness-aware async SGD (SASGD).

SASGD outperforms ASGD in terms of convergence, but it leaves performance on the table by treating all gradients as equal when computing staleness.
In this paper, we exploit that slack by modulating the learning rate as a function of moving averages of gradient statistics.
This idea yields better convergence performance, as we will show. 
We call this new algorithm Faster ASGD (FASGD for short).
We apply a similar idea to the problem of reducing bandwidth usage in distributed training contexts and achieve good results there as well.
We call this second algorithm Bandwidth-Aware Faster ASGD (B-FASGD for short).

The rest of the paper is organized as follows: first we discuss background and terminology.
We move on to cover the FASGD and B-FASGD algorithms.
Then we introduce our experimental architecture before finally finishing with the experimental results.

\section{The FASGD Algorithm}

We first introduce existing staleness aware algorithms before moving on to FASGD and variants.

\subsection{Background}
We adopt the notation from \cite{Suyog} for describing an asynchronous training session,
with some slight modifications:

\begin{itemize}
  \item $\lambda$: The number of clients.
  \item $\mu$: The minibatch size used by all learners to produce stochastic gradient estimates.
  \item $\alpha$: The master learning rate. Individual learning rates are made by modifications to this.
  \item $T$: The scalar timestamp that tracks the number of updates made to the master parameters.
    The timestamp starts at 0 and is incremented by one for each weight update (regardless
    of the number of clients or the sizes of the minibatches involved in computing the update). By the timestamp
    of a gradient, we mean the timestamp of the parameters used to compute that gradient.
  \item $\tau_{i,l}$: The step-staleness of the stochastic gradient from client $l$ with respect to the parameter timestamp
    $i$. A client $l$ pushes gradient with timestamp $j$ to the parameter server with timestamp $i$, where $i \geq j$.
    We calculate the step-staleness $\tau_{i,l}$ of this gradient as $i - j$.
    The step-staleness is always non-negative as defined.
\end{itemize}

Since there are many possible implementations of Asynchronous SGD, we formally specify the algorithm used for
the purposes of this paper:

\begin{itemize}
\item Async SGD Protocol: In this protocol, all $\lambda$ clients compute gradients asynchronously in parallel.
  When a client $l$ is done computing gradients $\Delta \theta^l$,
  it waits to take the lock on the parameter server (only one client can communicate with the
  server at a time). When the client takes the lock, the following things happen atomically before the lock is released:
  \begin{enumerate}
    \item Client $l$ passes the computed stochastic gradient $\Delta \theta^l$ to the server.
    \item The server updates the global parameters $\theta_i$ according to the following equation:
      $ \theta_{i + 1} = \theta_{i} - \alpha \Delta \theta^l$
    \item The server passes the updated parameters $\theta_{i + 1}$ back to client $l$. 
  \end{enumerate}

\end{itemize}

In \cite{Suyog}, they propose modulating the learning rate used on a per gradient basis.
Thus, their gradient update policy is as follows:

\begin{equation}
  g_i = \alpha(\tau_{i,k})\Delta \theta^k
\end{equation}

\begin{equation}
   \theta_{i + 1} = \theta_{i} - g_i
\end{equation}

where $\alpha(\tau_{i,k})$ is the step-staleness dependent learning rate
and $k$ is the client identifier.

They show that this modification allows convergence for larger values of $\lambda$.
This makes sense, because large values of $\lambda$ imply higher average step-staleness.

\subsection{Improvements}

One of the claims of this paper is that the step-staleness as defined in the last section
can be substantially improved upon as a measure of staleness.
In what follows, we introduce what we claim is a better measure. 

Consider a server with gradients $\theta_{i}$ and a client $l$ training on parameters $\theta_{j}$,
where $j \leq i$. Suppose that the client in question then sends a gradient estimate $\Delta \theta^l$
to the server with no intervening parameter updates.

We define the \textbf{B-Staleness} $\Gamma(\theta_{i}, \Delta \theta^l)$ as follows:

\begin{equation}  \label{bdef}
  \Gamma(\theta_{i}, \Delta \theta^l) = || (\Delta \theta^l) - (\Delta \theta_i) ||
\end{equation}

Where $ \Delta \theta^l $ is the stochastic gradient computed by client $l$ and
$ \Delta \theta_i$ is the stochastic gradient that would have been computed with client
$l$ on \textit{the same minibatch} if client $l$ had used $\theta_i$ instead of $\theta_j$.

It might not be immediately apparent why this is a better measure, but we'll justify this below.

The basic idea is as follows: consider a client $l$ training on parameters $\theta_{i-2}$
while the server has parameters $\theta_{i}$.
If client $l$ is the next client to send its stochastic gradient to the server, then that
gradient $\Delta \theta^l$ will have step staleness $\tau_{i,l} = 2$.
Now in this case the copy of the parameters on the master $\theta_i$ can be expressed as $\theta_i = \theta_{i-2} - g_1 - g_2$,
where $g_1$ and $g_2$ are the updates computed by multiplying some learning rate by the mean (across the minibatch) of
the the stochastic gradients returned by some clients. In the asynchronous context, $g_1$ and $g_2$ may have been
generated by clients with stale parameters as well.

However, not all pairs $g_1, g_2$ will cause the same amount of B-Staleness.
If $g_1$ and $g_2$ largely cancel each other out, then $\Gamma$ will be less than if $g_1$ and $g_2$ mostly have the same sign.
If the Hessian has large values at $\theta_i$, then $\Gamma$ will be higher than if the Hessian has small values.
It seems wise to account for these things in our synchronization protocol.

We care about staleness because the stochastic gradients computed using stale parameters will be a biased estimate
of the true gradient at the current point in parameter space.
In particular, for any stale gradient, we can decompose its difference from the true gradient
into the difference due to using SGD rather than batch GD and the difference $(\Delta \theta^l) - (\Delta \theta_i)$
(kind of a bias-variance decomposition for Async SGD.).
Since we can't control the first component (except by changing $\mu$),
it seems reasonable to say that, for our purposes, we just care about the B-Staleness.

B-Staleness as defined in equation \ref{bdef} can be approximated (using the Taylor series expansion)
by how far we have moved in the parameter space times the rate of change
in the gradients with respect to a change in parameter space.
This approximation is intractable to compute exactly because it involves higher order derivatives,
but to the extent we can approximate it,
we will be using a strictly more informative measure than in SASGD, and so should be able to improve convergence.

Taking inspiration from the version of RMSProp (introduced in \cite{RMSPROP}) in \cite{Graves}, we can do the following:
Maintain a global moving average of the standard deviation of each parameter.
At update time, modulate the learning rate per parameter based on this moving average and the step-staleness.
More formally, we propose the following protocol, which we call FASGD:

\begin{itemize}
\item FASGD: In this protocol, each learner $l$ pulls the weights from the parameter server,
  calculates the gradients and pushes the gradients to the parameter server,
  just as in the ASGD protocol defined earlier.
  As in ASGD, the parameter server updates
  the weights after receiving a gradient
  from any of the $\lambda$ learners.
  We maintain a moving average of gradient standard deviation as follows:

\begin{equation}
  n_i = \gamma n_{i-1} + (1 - \gamma) \Delta (\theta^l)^2
\end{equation}

\begin{equation}
   b_i = \gamma b_{i-1} + (1 - \gamma) \Delta \theta^l 
\end{equation}

\begin{equation}
   v_i = \beta  v_{i-1} + (1 - \beta)  \frac{1}{\sqrt{n_i - b_i^2 + \epsilon}}
\end{equation}
  
  where $\gamma$ and $\beta$ are hyperparameters and $\epsilon$ is for numerical
  stability. $n_i$ is the concatenation of all the parameters into a vector.
  $b_i$ and $v_i$ are the same shape, and all arithmetic (including the squaring)
  is elementwise.

  The FASGD weight update rule is then given by:

\begin{equation}
    g_i =  \frac{\alpha}{v_i \tau_{i,k}}\Delta \theta^k
\end{equation}

\begin{equation}
   \theta_{i + 1} = \theta_{i} - g_i 
\end{equation}

  where $k$ is the client identifier.  
  
\end{itemize}
It seems there ought to be a more principled relationship between the moving average
window and $\lambda$, but as shown in the Experimental Results section, we
are able to acheive good performance results with this heuristic method.

From an intutive perspective, it would be nice to understand why
dividing the learning rate by the standard deviation helps us.
One way to think of this is as follows:
If the Hessian has large values, we expect big changes from gradient to gradient,
so we should see high gradient standard deviation.
We also expect higher B-Staleness in this instance,
so dividing by a moving average of gradient standard deviation seems likely to help convergence.

We can also consider the potential for gradients to cancel each other out.
If gradients cancel each other out, they must have changed sign.
Where there are sign changes, there is likely to be higher gradient
standard deviation (or the gradients were small to begin with).
Thus, dividing by the standard deviation ought to account for cancellation effects.

Overall, FASGD should lead to better convergence than SASGD for the following two reasons:
First, it will allow us to keep the learning rate high when B-Staleness is less than step-staleness estimates.
Second, it will prevent us from overshooting when B-Staleness is higher than step-stalensss estimates.

\subsection{Bandwidth Aware FASGD}

In \cite{Dean}, they suggest reducing bandwidth consumption of distributed training algorithms
by limiting each model replica to request updated parameters only every $k_{fetch}$ steps and send updated gradient
values only every $k_{push}$ steps (where $k_{fetch}$ might not be equal to $k_{push}$).
This approach works to some extent, but it has two significant downsides.
First, the distribution of bandwidth consumption will be somewhat peaky.
This is suboptimal from a throughput perspective.
Second, the amount of achievable bandwidth reduction will depend on B-Staleness,
but this method fixes an amount at the beginning of training and sticks with it.

We attempt to fix both of those problems as follows.
First, we make the choice about whether to push or fetch at any given time a probabilistic choice.
That is, when a client has to decide whether to push or fetch, it generates a psuedo-random number and
compares that number to some other quantity.
If the pseudo-random number is greater than that quantity, the data is dropped - otherwise, it is transmitted.
Second, we modulate the likelihood of deciding to push or fetch at a given opportunity by the
moving average of gradient standard deviation already being computed for the FASGD algorithm.

More formally, given an opportunity to transmit a gradient or receive a parameter update,
a client will do so if and only if

\begin{equation} \label{req}
  r < \frac{1}{1 + \frac{c}{v + \epsilon}} 
\end{equation}

where $r \in [0,1] $ is a random number, $\epsilon$ is a small constant added for numerical stability,
$c$ is a hyper-parameter,
and v is the mean across all the parameters of the moving averages of their standard deviations as calculated by the FASGD server.
In practice, we will have separate hyper-parameters $c_{push}$ and $c_{fetch}$ for pushing and fetching.
We refer to the above algorithm (coupled with the FASGD policy) as the B-FASGD policy.

Note that if $v$ is very large, the right hand side of equation \ref{req} will be close to 1, so transmission is nearly assured.
If $v$ is close to 0 (it is always non-negative), the right hand side will be very small.
Thus the right hand side lies in $(0,1)$ and is inversely proportional to the gradient standard deviation.
This means that when we expect B-Staleness to be high, we will transmit more frequently,
incurring less step staleness.
When we expect B-Staleness to be low, we will skip more updates,
since each step missed corresponds to less accumulated B-Staleness.

It's worth mentioning that dropping parameters is different than dropping gradient updates.
When a client drops a parameter update, it simply continues computing gradients with the parameters it already has.
When a client decides to drop a gradient update, it's less clear what should be done.
We have elected to re-apply the most recent gradient from that client,
but this neccessitates maintaining a gradient cache on the server,
which could be prohibitive for a large values of $\lambda$ or large models.
It's possible that somehow averaging unsent gradients on the clients until transmission time would work better.
At any rate, it turns out (see the Experimental Results section), that dropping gradients is much less effective than dropping
parameters, so this decision may not matter in practice.

\section{Experimental Setup}
\label{sec:impl}
\vspace{-0.1mm}
In this section, we describe FRED - a Python library for simulating distributed training algorithms
deterministically on a single node.
The source code for FRED is at \href{https://github.com/DoctorTeeth/fasgd}{https://github.com/DoctorTeeth/fasgd}.
Later, we'll describe how we used FRED to generate experimental results for FASGD.

FRED works by taking an idiomatic description of a distributed training algorithm and running it
deterministically on a single node. This is useful for the following reasons:

First, determinism is helpful for debugging and experimentation.
It allows us to check that runs which should be bitwise equivalent are bitwise equivalent.
For instance, we can check that synchronous SGD with $\lambda$ clients and batch size $\mu$ is the same
algorithm as vanilla SGD with batch size $\frac{\mu}{\lambda}$.
We would have dramatically less confidence in the correctness of our implementations without this feature.

Second, setting up a big training cluster is an ordeal.
It's hard and it's expensive, and differences in configurations make
it hard to reproduce results.
Moreover, the quirks involved in a particular system may obscure algorithmic considerations.
Using a simulator allows you to test training algorithms for all topologies, since you
have full control over the order (which can be made probabilistic) in which updates are
sent from various clients.

To generate a FRED run, a user implements a Server class with a specific interface.
The implemented class governs how and when gradients will be applied to the Server Parameters.
More specifically, the Server class must implement an initialization function and an apply-update function.

The initialization function creates various data structures for keeping track of what's going on.
For instance, an FASGD server needs data structures to track gradient statistics.

The apply-update function takes 3 arguments: a gradient update, the client ID for the client that generated
the update, and the time stamp of the parameters used by the client to generate the update.
Then the update is applied to the parameters in the appropriate way.
For instance, in a synchronous SGD server, the apply update function is implemented with the
following Python code:

\begin{verbatim}

def apply_update(self, grads, timestamp, client):

        unblock = False
        self.pending_grads[client] = grads

        if len(self.pending_grads) == self.clients:

            # all grads is a list of lists of grads
            # each of which has the same length
            all_grads = self.pending_grads.values()

            # apply the param update
            for this_grad in all_grads:
                for g, p in zip(this_grad, self.params):
                    old_p = p.get_value()
                    mod = g / self.clients 
                    p.set_value(old_p - self.learning_rate * mod)

            self.timestamp += 1 # weights have changed
            unblock = True
            self.pending_grads = {}

        return self.params, self.timestamp, unblock
\end{verbatim}

Specifying the behavior of those two functions uniquely defines the behavior of a
FRED Server.

The Server is then hooked up to a Dispatcher and a group of Clients.
The Dispatcher will manage the simulation of which gradient update come from what client when.
This simulation takes as an argument a rule determining each client's probability of being selected
and how that probability will change upon that client having been selected.
The dispatcher will also take as arguments functions that determine whether and how it will fetch and/or push
parameters and gradients, respectively.

We've made at least one opinionated decision about the library that merits further discussion.
After a client pushes gradients to a server, it will wait for the resulting parameter update (or non-update) before
resuming its computation. This is good in some ways and bad in others. If the overhead of applying the updates is
small relative to that of computing the gradients, and/or if we care a lot about staleness, we may as well wait.
If we cared a lot about throughput instead, we might just work with whatever the most recently computed set of parameters is.
We may decide to make this configurable, or to come up with some more general framework in which the configurability
of this particular decision comes for free.

\vspace{-0.1mm}

\section{Experimental Results}
\label{expts}

In what follows, we discuss experimental results generated using FRED.
We compare FASGD to SASGD on the MNIST task from \cite{MNIST} in a variety of contexts.
We also compare FASGD to B-FASGD.

\subsection{FASGD}

We test the FASGD protocol using a 2 layer MLP trained on the MNIST task.
The MLP has 200 hidden units.
These use a relu activation, and the cost is negative log likelihood,
as is standard.

\begin{figure}[ht!]
  \centering
  \begin{minipage}[b]{0.5\linewidth}
    \includegraphics[scale=0.35]{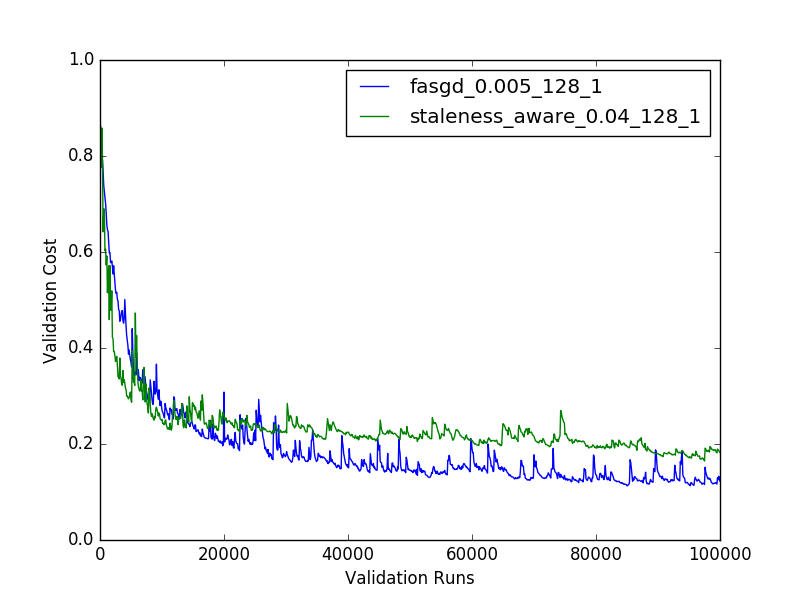}
  \end{minipage}
  \begin{minipage}[b]{0.5\linewidth}
    \includegraphics[scale=0.35]{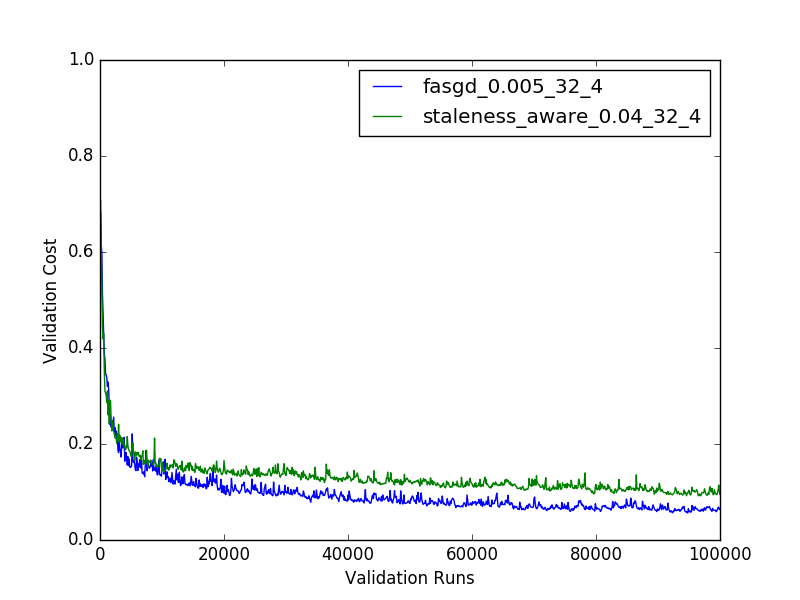}
  \end{minipage} 
  \begin{minipage}[b]{0.5\linewidth}
    \includegraphics[scale=0.35]{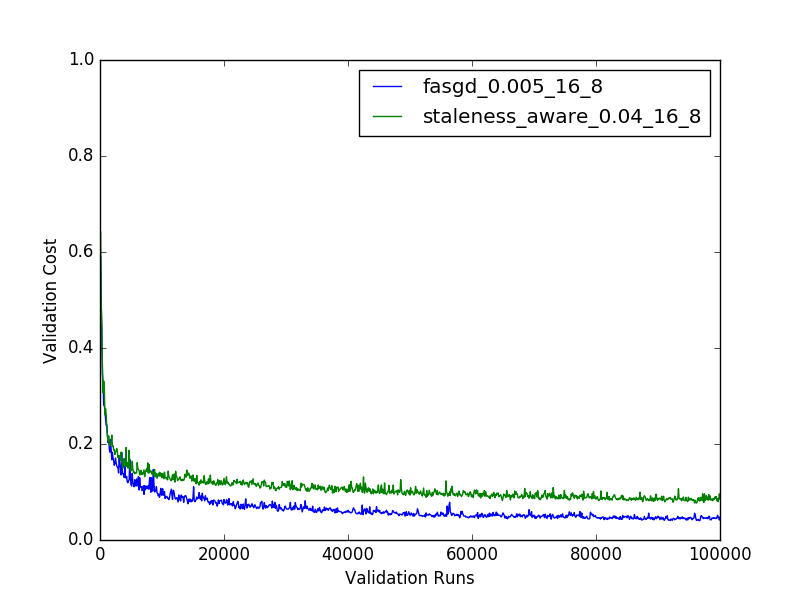}
  \end{minipage}
  \begin{minipage}[b]{0.5\linewidth}
    \includegraphics[scale=0.35]{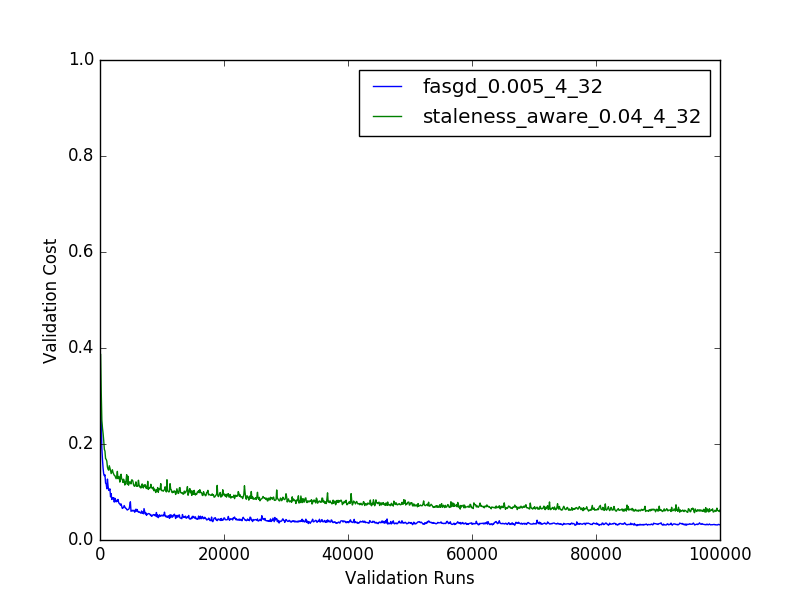}
  \end{minipage}
  \caption{MNIST Validation cost for FASGD (Blue) and SASGD (Green).
  FASGD outperforms SASGD for all tested combinations of $\mu$ and $\lambda$. 
  Left to right and top to bottom, the batch sizes are 1, 4, 8, and 32.
  The product of $\mu$ and $\lambda$ is held constant at 128.
  }
  \label{costfig} 
\end{figure}

Results are shown in figure \ref{costfig}, and are generated in the following way:
We considered 4 combinations of $\mu$ and $\lambda$: namely $\mu=1, \lambda=128$, 
$\mu=4, \lambda=32$, $\mu=8, \lambda=16$, and $\mu=32, \lambda=4$.
The product $\mu \lambda$ was kept constant in order to keep convergence performance
relatively similar across all 4 runs as advised in \cite{Tradeoffs}.
We separately choose the best learning rate (across the set of 4 combinations)
for each of FASGD and SASGD from a pool of 16 candidate learning rates.
That rate was 0.005 for FASGD and 0.04 for SASGD, as can be seen from the labels in the figure.
Each configuration ran for 100,000 iterations, 
where each iteration corresponds to a client calclulating
a gradient estimate for a single minibatch.
FASGD performs meaningfully better (converging faster and to a better cost)
regardless of $\mu$ and $\lambda$.
This difference is also robust with respect to different random
initializations.
For the sake of fairness, we did not tune the FASGD hyper-parameters.
Had we done so, we expect that FASGD would have outperformed SASGD by
a larger margin.

Some of these $\mu$ values might seem small, but they were chosen for comparability
with experiments from \cite{Suyog}. Moreover, work by \cite{winograd} suggests that, at least
for convolutional networks, good throughput should be achievable for minibatch sizes between
1 and 64. It's also worth mentioning that the next experiment shows FASGD outperforming SASGD for $\mu = 128$,
which is a completely standard minibatch size.

\begin{figure}[ht!]
  \centering
  \begin{minipage}[b]{0.5\linewidth}
    \includegraphics[scale=0.35]{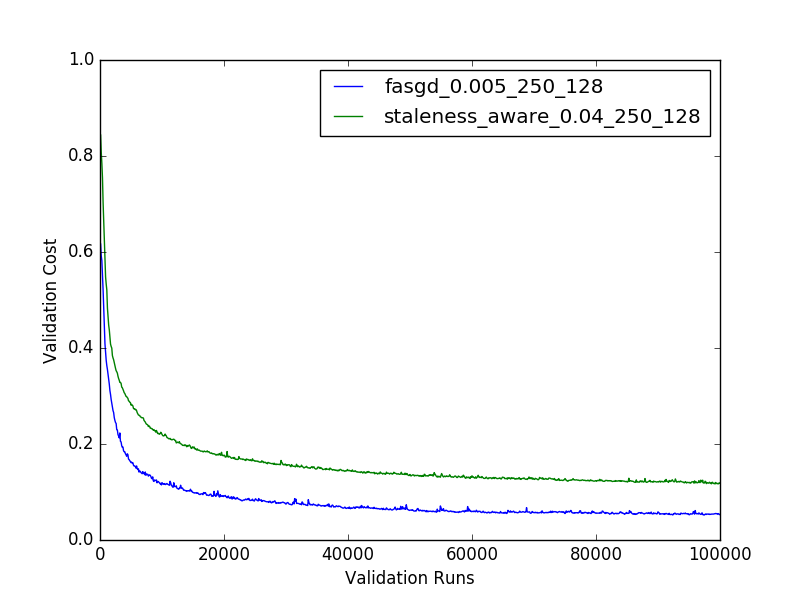}
  \end{minipage}
  \begin{minipage}[b]{0.5\linewidth}
    \includegraphics[scale=0.35]{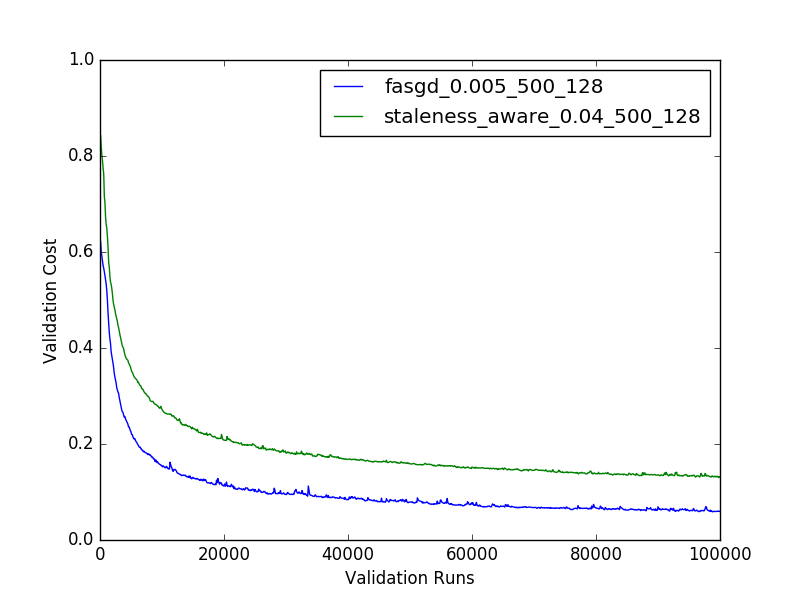}
  \end{minipage} 
  \begin{minipage}[b]{0.5\linewidth}
    \includegraphics[scale=0.35]{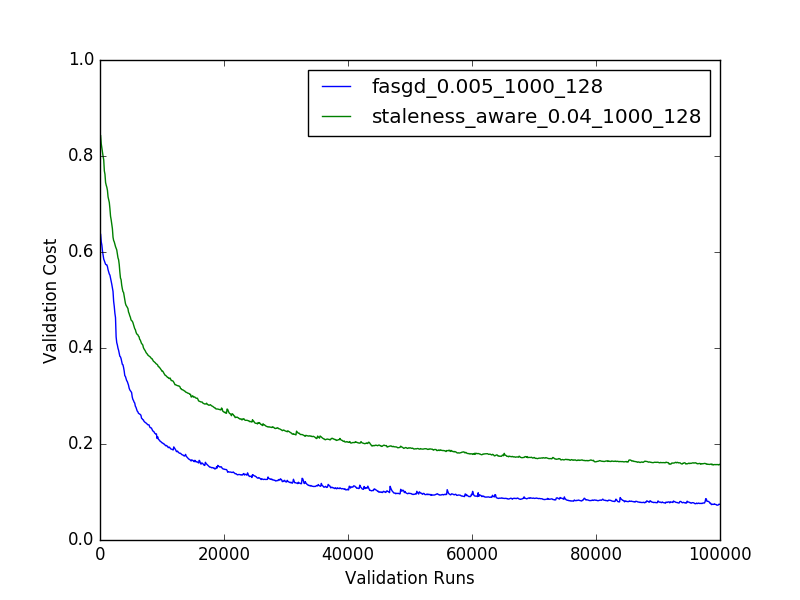}
  \end{minipage}
  \begin{minipage}[b]{0.5\linewidth}
    \includegraphics[scale=0.35]{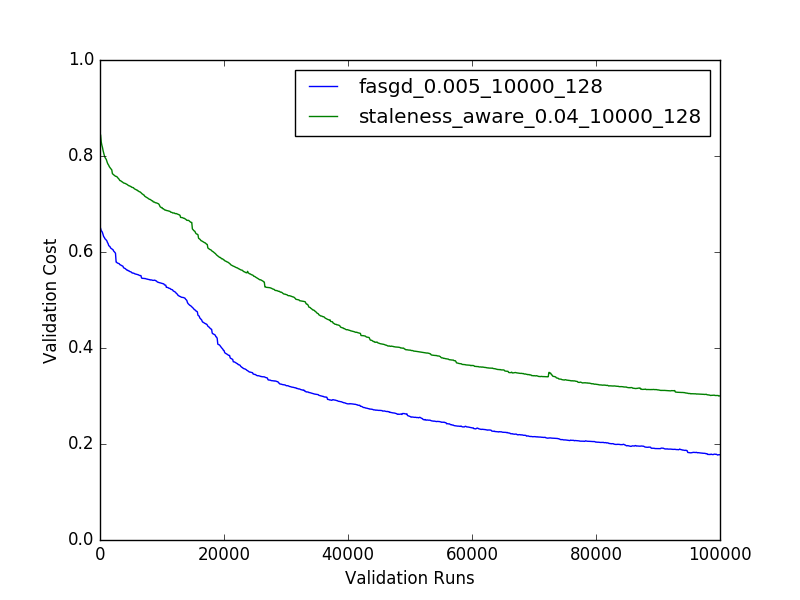}
  \end{minipage}
  \caption{MNIST Validation cost for FASGD (Blue) and SASGD (Green),
    tested with different $\lambda$ values.
    Left to right and top to bottom, the $\lambda$ values used are 250, 500,
    1000, 10000.
  }
  \label{clientfig} 
\end{figure}

We also tested how the difference between FASGD and SASGD changed as a function of $\lambda$.
The results are displayed in figure \ref{clientfig}.
We tried $\lambda$ values of 250, 500, 1000, and 10,000.
This was with $\mu = 128$ and the same learning rates from the first experiment.
We used the ASGD policy described earlier.
FASGD beats SASGD in all instances, but the relative outperformance increases as $\lambda$ goes up.
Since overall staleness increases as $\lambda$ goes up, this provides evidence in favor of our hypothesis
that FASGD helps more when staleness is higher.
These results also suggest that FASGD might allow improved speed-up in distributed training contexts
by helping us to increase $\lambda$.

\subsection{Bandwidth Aware FASGD}

We use the same model from before, also on MNIST.
In this section, however, we are interested in how much we can reduce bandwidth
usage while still achieving good convergence performance.

\begin{figure}[ht!]
  \centering
  \begin{minipage}[b]{0.5\linewidth}
    \includegraphics[scale=0.35]{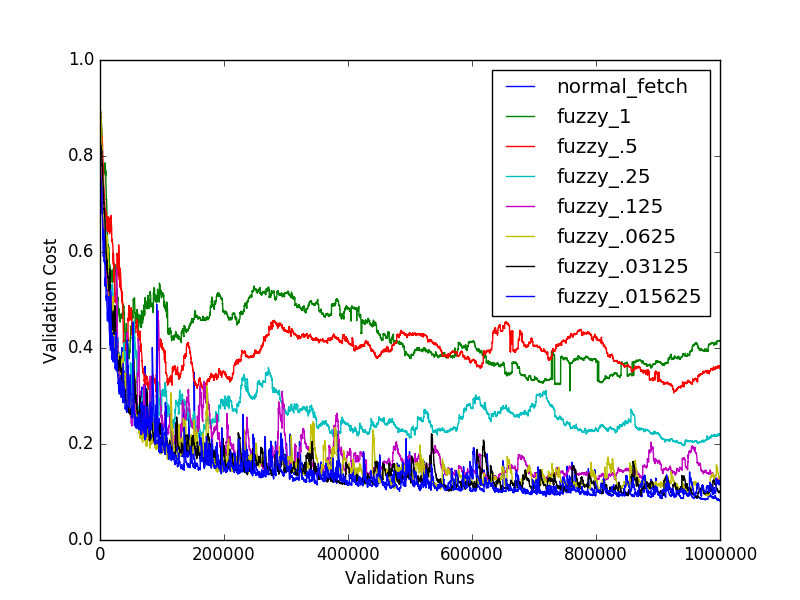}
  \end{minipage}
  \begin{minipage}[b]{0.5\linewidth}
    \includegraphics[scale=0.35]{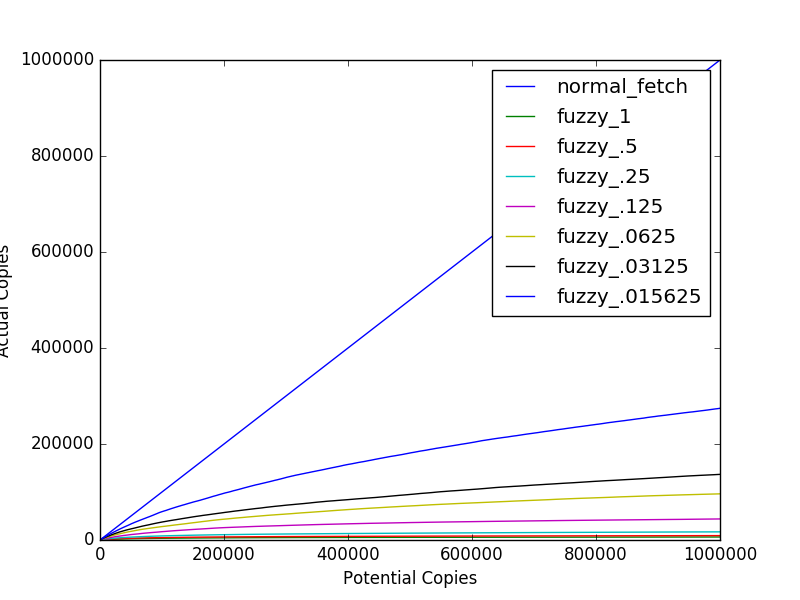}
  \end{minipage} 
  \begin{minipage}[b]{0.5\linewidth}
    \includegraphics[scale=0.35]{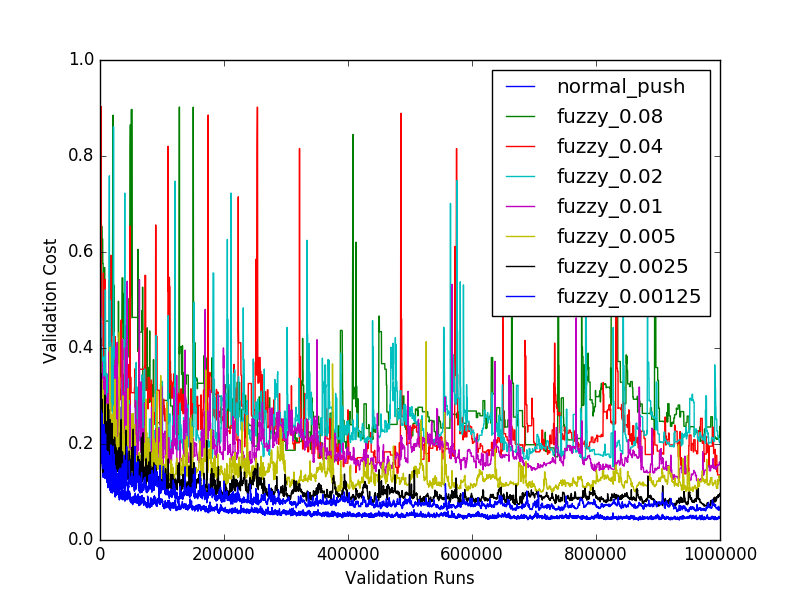}
  \end{minipage}
  \begin{minipage}[b]{0.5\linewidth}
    \includegraphics[scale=0.35]{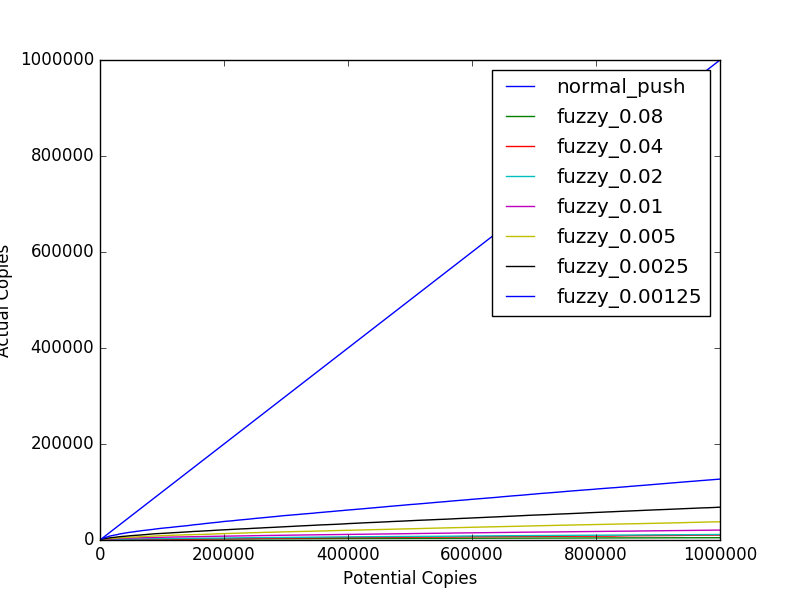}
  \end{minipage}
  \caption{
    Convergence and bandwidth usage (both push and fetch)
    of B-FASGD for a variety of c-values.
    The top row shows the results of modulating just kfetch.
    The bottom row shows the results of modulating just kpush.
  }
  \label{bandwidthfig} 
\end{figure}

We use the B-FASGD protocol defined earlier with the standard FASGD model as a baseline - we
show results in figure \ref{bandwidthfig}.

As above, we divide bandwidth usage into copies of the parameters from the server to a client and
copies of the gradients from a client to the server (fetches and pushes, respectively).
We find that fetch usage can be reduced significantly without seriously impacting performance,
while reducing push usage quickly causes training to diverge.

In particular, the experiments show that it's possible to achieve reduction of a factor of
10 in fetch communication (corresponding to a reduction of a factor of 5 in total bandwidth use),
but that attempts to reduce push communication even a small amount were less successful.
It's possible that this is an artifact of our simplistic method for dropping gradient updates,
and that some more sophisticated method would do better.

One thing that merits special attention is that the graphs of copies versus potential
copies have negative `second derivatives'. This suggests that
it makes sense to modulate bandwidth usage by gradient statistics.
If the system is bottlenecked by bandwidth early in training, it doesn't need to be
bottlenecked later in training.
It also suggests an interesting practical application - more clients could
be dynamically provisioned as training progresses, keeping bandwidth usage steady.

\vspace{-2.5mm}
\section{Future Work}
\vspace{-1mm}

We are excited to explore further extensions to this work.

We would like to expand B-FASGD by synchronizing parameters on a per-tensor basis.
Since we are dynamically choosing when to synchronize using per-parameter moving averages,
we could choose to update parameters more frequently when their moving averages
indicate a higher likelihood of staleness, and vice versa.

It could potentially be interesting to fix a bandwidth budget and use the gradient
statistics to dynamically allocate portions of that budget to different tensors
according to likelihood of staleness.

We would also like to explore why convergence performance is so much more sensitive
to changes in the push rate than it is to changes in the fetch rate.

Finally, it would be nice to give convergence gurarantees of the form in \cite{Lian} for FASGD.
We would further like to more formally characterize the trade-off between bandwidth usage and convergence,
and the trade-off between accuracy of modulation and computation cost.
These last two strike us as fundamental questions worth more consideration.
There may be some interplay between this and formal work in distributed systems.

\vspace{-2.5mm}
\section{Conclusion}
\vspace{-1mm}

In summary, the key technical contributions of this work are as follows:

\begin{itemize}

\item We introduce a novel algorithm (FASGD) for asynchronous gradient descent that converges faster and to lower cost than
the current state of the art. 

\item We demonstrate that an extension of these ideas can be used to significantly reduce bandwidth costs in a distributed context.

\item We provide an open source framework for evaluating distributed training algorithms in a deterministic way on a single machine.

\end{itemize}

We hope that this work will seriously accelerate distributed training of large, deep neural networks in
all contexts,
but we expect it to be especially important in contexts where the staleness distribution is poorly behaved.
For instance, when the training cluster is large and heterogenous, we expect FASGD to outperform SASGD even more.
FASGD could also be useful for distributed training of conditional networks as in \cite{Conditional}.
If only certain tensors are updated for a given input,
it may be beneficial to selectively send per-tensor updates over the network. 

{\bf Acknowledgments}: We thank the developers of Theano (\cite{Theano}).
We also thank Yoshua Bengio for helpful discussions and feedback.

\newpage

\nocite{*}
\bibliography{fasgd}
\bibliographystyle{fasgd}

\end{document}